\documentclass{llncs}
\usepackage{epsf}

\usepackage[utf8]{inputenc}
\usepackage{times}
\usepackage{latexsym}
 
\usepackage{textcomp}
 
\usepackage{multirow}
 
\usepackage{url}
\usepackage{amsmath,amsfonts,amssymb}
\usepackage{textcomp}
 
\usepackage{pbox}
\usepackage{graphicx}
 
\usepackage{multicol}
 
\usepackage{url}
 
\begin{document}
 
\title{Generating~Word~and~Document~Embeddings
for~Sentiment~Analysis}
 
\author{
Cem Rıfkı Aydın, Tunga Güngör and Ali Erkan}
\institute{Boğaziçi University \\
Computer Engineering Department\\
Bebek, Istanbul, Turkey, 34342\\
cemrifkiaydin@gmail.com, gungort@boun.edu.tr, alierkan@gmail.com}
 
\maketitle
 
\begin{abstract}
 
Sentiments of words differ from one corpus to another. Inducing general sentiment lexicons for languages and using them cannot, in general, produce meaningful results for different domains. In this paper, we combine contextual and supervised information with the general semantic representations of words occurring in the dictionary. Contexts of words help us capture the domain-specific information and supervised scores of words are indicative of the polarities of those words. When we combine supervised features of words with the features extracted from their dictionary definitions, we observe an increase in the success rates. We try out the combinations of contextual, supervised, and dictionary-based approaches, and generate original vectors. We also combine the word2vec approach with hand-crafted features. We induce domain-specific sentimental vectors for two corpora, which are the movie domain and the Twitter datasets in Turkish. When we thereafter generate document vectors and employ the support vector machines method utilising those vectors, our approaches perform better than the baseline studies for Turkish with a significant margin. We evaluated our models on two English corpora as well and these also outperformed the word2vec approach. It shows that our approaches are cross-domain and portable to other languages.
 
\hfill
 
\textbf{Keywords: }Sentiment Analysis, Opinion Mining, Word Embeddings, Machine Learning
 
\end{abstract}
 
\section{Introduction}
 
Sentiment analysis has recently been one of the hottest topics in natural language processing (NLP). It is used to identify and categorise opinions expressed by reviewers on a topic or an entity. Sentiment analysis can be leveraged in marketing, social media analysis, and customer service. Although many studies have been conducted for sentiment analysis in widely spoken languages, this topic is still immature for Turkish and many other languages.
 
Neural networks outperform the conventional machine learning algorithms in most classification tasks, including sentiment analysis~\cite{Goldberg:16}. In these networks, word embedding vectors are fed as input to overcome the data sparsity problem and to make the representations of words more “meaningful” and robust. Those embeddings indicate how close the words are to each other in the vector space model (VSM).
 
Most of the studies utilise embeddings, such as word2vec~\cite{Mikolov:13}, which take into account the syntactic and semantic representations of the words only. Discarding the sentimental aspects of words may lead to words of different polarities being close to each other in the VSM, if they share similar semantic and syntactic features.
 
For Turkish, there are only a few studies which leverage sentimental information in generating the word and document embeddings. Unlike the studies conducted for English and other widely-spoken languages, in this paper, we use the official dictionaries for this language and combine the unsupervised and supervised scores to generate a unified score for each dimension of the word embeddings in this task.
 
Our main contribution is to create original and effective word vectors that capture syntactic, semantic and sentimental characteristics of words, and use all of this knowledge in generating embeddings. We also utilise the word2vec embeddings trained on a large corpus. Besides using these word embeddings, we also generate hand-crafted features on a review-basis and create document vectors. We evaluate those embeddings on two datasets. The results show that we outperform the approaches which do not take into account the sentimental information. We also had better performances than other studies carried out on sentiment analysis in Turkish media. We also evaluated our novel embedding approaches on two English corpora of different genres. We outperformed the baseline approaches for this language as well. The source code and datasets are publicly available\footnote{\url{https://github.com/cemrifki/sentiment-embeddings}}.
 
The paper is organised as follows. In Section 2, we present the existing works on sentiment classification. In Section 3, we describe the methods proposed in this work. The experimental results are shown and the main contributions of our proposed approach are discussed in Section 4. In Section 5, we conclude the paper.

\section{Related Work}
In the literature, the main consensus is that the use of dense word embeddings outperforms the sparse embeddings in many tasks. Latent semantic analysis (LSA) used to be the most popular method in generating word embeddings before the invention of the word2vec and other word vector algorithms which are mostly created by shallow neural network models. Although many studies have been employed on generating word vectors including both semantic and sentimental components, generating and analysing the effects of different types of embeddings on different tasks is an emerging field for Turkish.
 
Latent Dirichlet allocation (LDA) is used in \cite{Blei:03} to extract mixture of latent topics. However, it focusses on finding the latent topics of a document, not the word meanings themselves. In \cite{Turney:10}, LSA is utilised to generate word vectors, leveraging indirect cooccurrence statistics. These outperform the use of sparse vectors \cite{Cortes:95}.
 
Some of the prior studies have also taken into account the sentimental characteristics of a word when creating word vectors \cite{Li:10,Lin:09,Boyd-Graber:10}. A model with semantic and sentiment components is built in \cite{Maas:11}, making use of star-ratings of reviews. In \cite{Hamilton:16}, a sentiment lexicon is induced preferring the use of domain-specific cooccurrence statistics over the word2vec method and they outperform the latter.
 
In a recent work on sentiment analysis in Turkish \cite{Ert:17}, they learn embeddings using Turkish social media. They use the word2vec algorithm, create several unsupervised hand-crafted features, generate document vectors and feed them as input into the support vector machines (SVM) approach. We outperform this baseline approach using more effective word embeddings and supervised hand-crafted features.
 
In English, much of the recent work on learning sentiment-specific embeddings relies only on distant supervision. In \cite{Fel:17}, emojis are used as features and a bi-LSTM (bi-directional long short-term memory) neural network model is built to learn sentiment-aware word embeddings. In \cite{Tan:14}, a neural network that learns word embeddings is built by using contextual information about the data and supervised scores of the words. This work captures the supervised information by utilising emoticons as features. Most of our approaches do not rely on a neural network model in learning embeddings. However, they produce state-of-the-art results.
 
\section{Methodology}
 
We generate several word vectors, which capture the sentimental, lexical, and contextual characteristics of words. In addition to these mostly original vectors, we also create word2vec embeddings to represent the corpus words by training the embedding model on these datasets. After generating these, we combine them with hand-crafted features to create document vectors and perform classification, as will be explained in Section 3.5.
 
\subsection{Corpus-based Approach}
Contextual information is informative in the sense that, in general, similar words tend to appear in the same contexts. For example, the word \emph{smart} is more likely to cooccur with the word \emph{hardworking} than with the word \emph{lazy}. This similarity can be defined semantically and sentimentally. In the corpus-based approach, we capture both of these characteristics and generate word embeddings specific to a domain.
 
Firstly, we construct a matrix whose entries correspond to the number of cooccurrences of the row and column words in sliding windows. Diagonal entries are assigned the number of sliding windows that the corresponding row word appears in the whole corpus. We then normalise each row by dividing entries in the row by the maximum score in it.
 
Secondly, we perform the principal component analysis (PCA) method to reduce the dimensionality. It captures latent meanings and takes into account high-order cooccurrence removing noise. The attribute (column) number of the matrix is reduced to 200. We then compute cosine similarity between each row pair $w_i$ and $w_j$ as in \eqref{eq:1} to find out how similar two word vectors (rows) are.
 
\begin{align}
    \boldsymbol{cos}(w_{i}, w_{j}) = \frac{w_{i}^\intercal \dot w_{j}}{||w_i|| \dot ||w_j||} \label{eq:1}
\end{align}
 
Thirdly, all the values in the matrix are subtracted from 1 to create a dissimilarity matrix. Then, we feed the matrix as input into the fuzzy c-means clustering algorithm. We chose the number of clusters as 200, as it is considered a standard for word embeddings in the literature. After clustering, the dimension \emph{i} for a corresponding word indicates the degree to which this word belongs to cluster \emph{i}. The intuition behind this idea is that if two words are similar in the VSM, they are more likely to belong to the same clusters with analogous probabilities. In the end, each word in the corpus is represented by a 200-dimensional vector.
 
In addition to this method, we also perform singular value decomposition (SVD) on the cooccurrence matrices, where we compute the matrix $\boldsymbol{M}^{PPMI} = \boldsymbol{U\Sigma} V^{T}$. Positive pointwise mutual information (PPMI) scores between words are calculated and the truncated singular value decomposition is computed. We take into account the \textit{\textbf{U}} matrix only for each word. We have chosen the singular value number as 200. That is, each word in the corpus is represented by a 200-dimensional vector as follows.

\begin{align}
    \boldsymbol{w}_i = (\boldsymbol{U})_i \label{eq:2}
\end{align}
 
\subsection{Dictionary-based Approach}
In Turkish, there do not exist well-established sentiment lexicons as in English. In this approach, we made use of the TDK (T\"{u}rk Dil Kurumu - “Turkish Language Institution”) dictionary to obtain word polarities. Although it is not a sentiment lexicon, combining it with domain-specific polarity scores obtained from the corpus led us to have state-of-the-art results.
 
We first construct a matrix whose row entries are corpus words and column entries are the words in their dictionary definitions. We followed the Boolean approach. For instance, for the word \emph{cat}, the column words occurring in its dictionary definition are given a score of 1. Those column words not appearing in the definition of \emph{cat} are assigned a score of 0 for that corresponding row entry.
 
When we performed clustering on this matrix, we observed that those words having similar meanings are, in general, assigned to the same clusters. However, this similarity fails in capturing the sentimental characteristics. For instance, the words \emph{happy} and \emph{unhappy} are assigned to the same cluster, since they have the same words, such as \emph{feeling}, in their dictionary definitions. However, they are of opposite polarities and should be discerned from each other.
 
Therefore, we utilise a metric to move such words away from each other in the VSM, even though they have common words in their dictionary definitions. We multiply each value in a row with the corresponding row word's raw supervised score, thereby having more meaningful clusters. Using the training data only, the supervised polarity score per word is calculated as in \eqref{eq:2}.
 
\begin{align}
w_{t} &=\log\frac{\frac{N_{t}}{N}+0.01}{\frac{N'_{t}}{N'}+0.01} \label{eq:2}
\end{align}
 
Here, $ w_{t}$ denotes the sentiment score of word $t$, $N_{t}$ is the number of documents (reviews or tweets) in which $t$ occurs in the dataset of positive polarity, $N$ is the number of all the words in the corpus of positive polarity. $N'$ denotes the corpus of negative polarity. $N'_{t}$ and $N'$ denote similar values for the negative polarity corpus. We perform normalisation to prevent the imbalance problem and add a small number to both numerator and denominator for smoothing.
 
As an alternative to multiplying with the supervised polarity scores, we also separately multiplied all the row scores with only +1 if the row word is a positive word, and with -1 if it is a negative word. We have observed it boosts the performance more compared to using raw scores.
 
\begin{figure}[!htpb]
\begin{center}
\vspace{0.0cm}
\includegraphics[width=5.45cm]{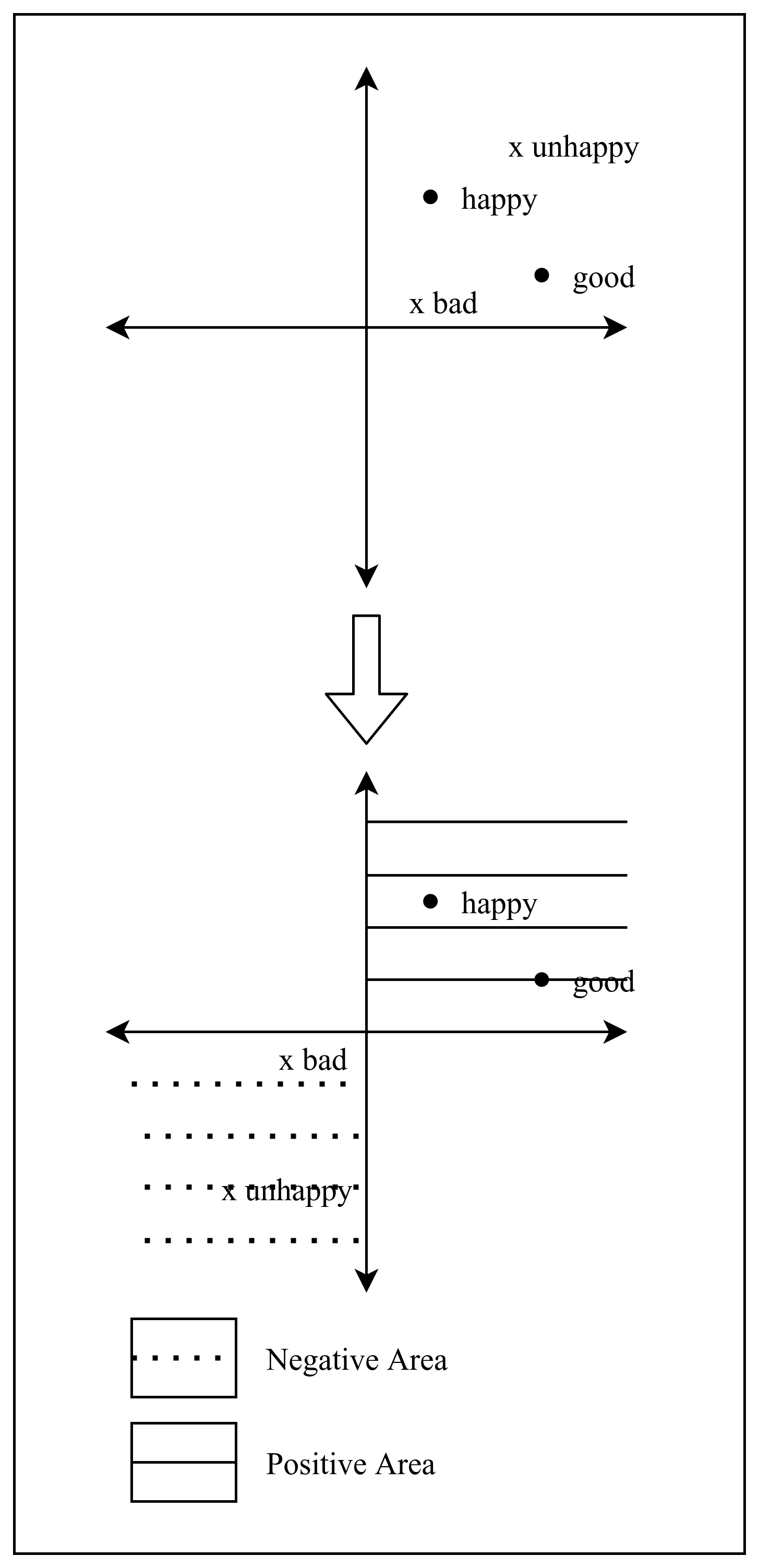}
\caption{The effect of using the supervised scores of words in the dictionary algorithm. It shows how sentimentally similar word vectors get closer to each other in the VSM.}
\label{supervDict}
 
\setlength{\belowdisplayskip}{0.5pt}
\end{center}
\end{figure}

The effect of this multiplication is exemplified in Figure \ref{supervDict}, showing the positions of word vectors in the VSM. Those ``x" words are sentimentally negative words, those ``o" words are sentimentally positive ones. On the top coordinate plane, the words of opposite polarities are found to be close to each other, since they have common words in their dictionary definitions. Only the information concerned with the dictionary definitions are used there, discarding the polarity scores. However, when we utilise the supervised score (+1 or -1), words of opposite polarities (e.g. ``happy" and ``unhappy") get far away from each other as they are translated across coordinate regions. Positive words now appear in quadrant 1, whereas negative words appear in quadrant 3. Thus, in the VSM, words that are sentimentally similar to each other could be clustered more accurately. Besides clustering, we also employed the SVD method to perform dimensionality reduction on the unsupervised dictionary algorithm and used the newly generated matrix by combining it with other subapproaches. The number of dimensions is chosen as 200 again according to the $U$ matrix. The details are given in Section 3.4. When using and evaluating this subapproach on the English corpora, we used the SentiWordNet lexicon~\cite{Bac:10}. We have achieved better results for the dictionary-based algorithm when we employed the SVD reduction method compared to the use of clustering.
 
\subsection{Supervised Contextual 4-scores}
Our last component is a simple metric that uses four supervised scores for each word in the corpus. We extract these scores as follows. For a target word in the corpus, we scan through all of its contexts. In addition to the target word's polarity score (the self-score), out of all the polarity scores of words occurring in the same contexts as the target word, minimum, maximum, and average scores are taken into consideration. The word polarity scores are computed using \eqref{eq:2}. Here, we obtain those scores from the training data.
 
The intuition behind this method is that those four scores are more indicative of a word's polarity rather than only one (the self-score). This approach is fully supervised unlike the previous two approaches.
 
\subsection{Combination of the Word Embeddings}
In addition to using the three approaches independently, we also combined all the matrices generated in the previous approaches. That is, we concatenate the reduced forms (SVD - U) of corpus-based, dictionary-based, and the whole of 4-score vectors of each word, horizontally. Accordingly, each corpus word is represented by a 404-dimensional vector, since corpus-based and dictionary-based vector components are each composed of 200 dimensions, whereas the 4-score vector component is formed by four values.
 
The main intuition behind the ensemble method is that some approaches compensate for what the others may lack. For example, the corpus-based approach captures the domain-specific, semantic, and syntactic characteristics. On the other hand, the 4-scores method captures supervised features, and the dictionary-based approach is helpful in capturing the general semantic characteristics. That is, combining those three approaches makes word vectors more representative.
 
\subsection{Generating Document Vectors}
 
After creating several embeddings as mentioned above, we create document (review or tweet) vectors. For each document, we sum all the vectors of words occurring in that document and take their average. In addition to it, we extract three hand-crafted polarity scores, which are minimum, mean, and maximum polarity scores, from each review. These polarity scores of words are computed as in \eqref{eq:2}. For example, if a review consists of five words, it would have five polarity scores and we utilise only three of these sentiment scores as mentioned. Lastly, we concatenate these three scores to the averaged word vector per review.
 
\begin{figure}[h]
\begin{center}
\includegraphics[width=10.8cm]{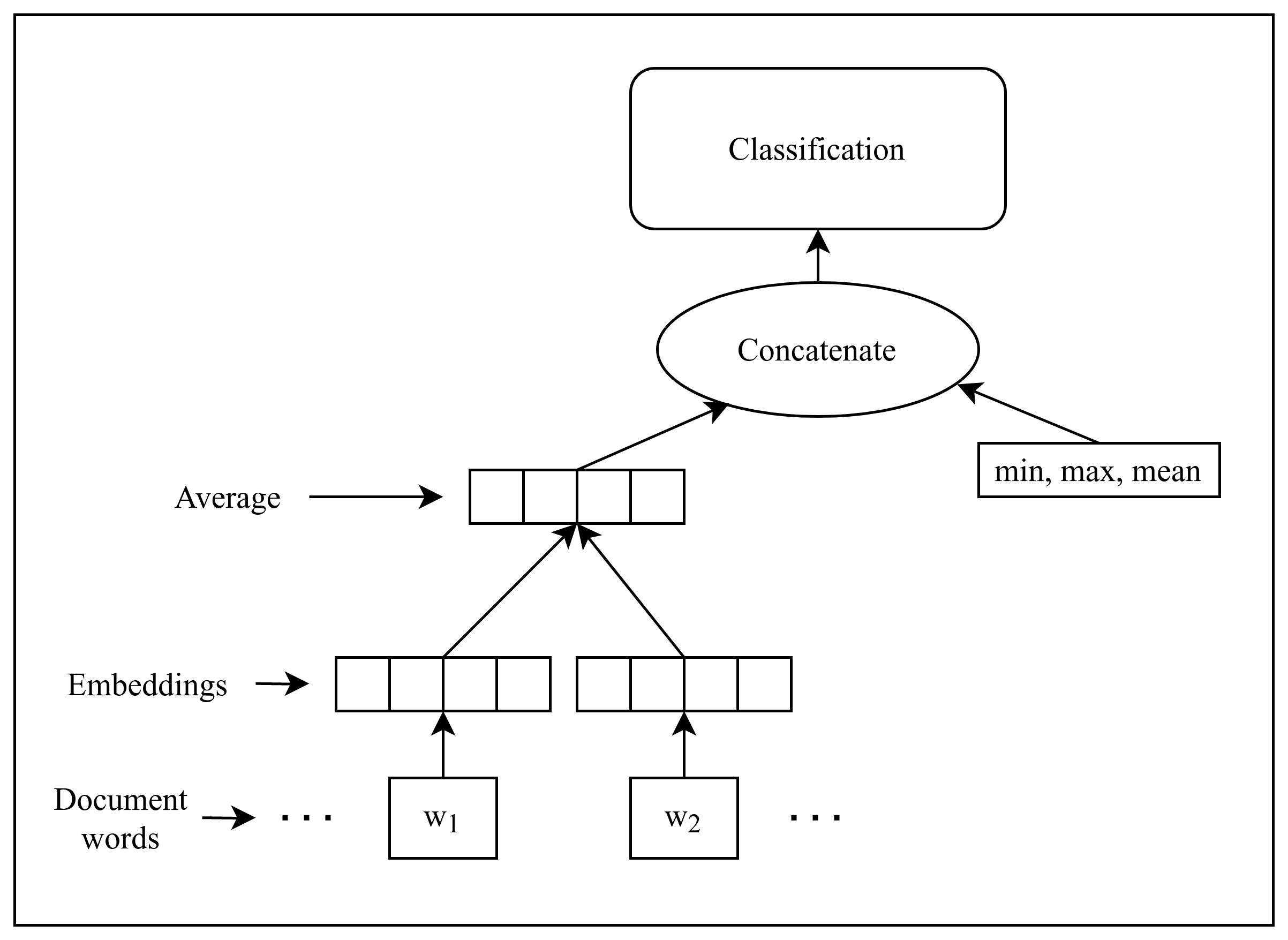}
\caption{The flowchart of our system.}
\label{classify}
\end{center}
\end{figure}

That is, each review is represented by the average word vector of its constituent word embeddings and three supervised scores. We then feed these inputs into the SVM approach. The flowchart of our framework is given in Figure \ref{classify}. When combining the unsupervised features, which are word vectors created on a word-basis, with supervised three scores extracted on a review-basis, we have better state-of-the-art results.
 
\section{Datasets}
 
We utilised two datasets for both Turkish and English to evaluate our methods.
 
For Turkish, as the first dataset, we utilised the movie reviews which are collected from a popular website\footnote{\url{https://www.beyazperde.com}}. The number of reviews in this movie corpus is 20,244 and the average number of words in reviews is 39. Each of these reviews has a star-rating score which is indicative of sentiment. These polarity scores are between the values 0.5 and 5, at intervals of 0.5. We consider a review to be negative it the score is equal to or lower than 2.5. On the other hand, if it is equal to or higher than 4, it is assumed to be positive. We have randomly selected 7,020 negative and 7,020 positive reviews and processed only them.
 
The second Turkish dataset is the Twitter corpus which is formed of tweets about Turkish mobile network operators. Those tweets are mostly much noisier and shorter compared to the reviews in the movie corpus. In total, there are 1,716 tweets. 973 of them are negative and 743 of them are positive. These tweets are manually annotated by two humans, where the labels are either positive or negative. We measured the Cohen's Kappa inter-annotator agreement score to be 0.82. If there was a disagreement on the polarity of a tweet, we removed it.
 
We also utilised two other datasets in English to test the portability of our approaches to other languages. One of them is a movie corpus collected from the web\footnote{\url{https://github.com/dennybritz/cnn-text-classification-tf}}. There are 5,331 positive reviews and 5,331 negative reviews in this corpus. The other is a Twitter dataset, which has nearly 1.6 million tweets annotated through a distant supervised method~\cite{Go:09}. These tweets have positive, neutral, and negative labels. We have selected 7,020 positive tweets and 7,020 negative tweets randomly to generate a balanced dataset.
 
\section{Experiments}
\subsection{Preprocessing}
In Turkish, people sometimes prefer to spell English characters for the corresponding Turkish characters (e.g. \emph{i} for \emph{\i}, \emph{c} for \emph{\c{c}}) when writing in electronic format. To normalise such words, we used the Zemberek tool \cite{Akin:07}. All punctuation marks except ``!" and ``?" are removed, since they do not contribute much to the polarity of a document. We took into account emoticons, such as ``:))", and idioms, such as ``kafay{\i} yemek'' (lose one's mind), since two or more words can express a sentiment together, irrespective of the individual words thereof. Since Turkish is an agglutinative language, we used the morphological parser and disambiguation tools \cite{Sak:07,Sak:08}. We also performed negation handling and stop-word elimination. In negation handling, we append an underscore to the end of a word if it is negated. For example, ``güzel değil" (not beautiful) is redefined as ``güzel\_" (beautiful\_) in the feature selection stage when supervised scores are being computed.
 
\subsection{Hyperparameters}
 
We used the LibSVM utility of the WEKA tool. We chose the linear kernel option to classify the reviews. We trained word2vec embeddings on all the four corpora using the Gensim library~\cite{Rad:10} with the skip-gram method. The dimension size of these embeddings is set at 200. As mentioned, other embeddings, which are generated utilising the clustering and the SVD approach, are also of size 200. For c-means clustering, we set the maximum number of iterations at 25, unless it converges.
 
\subsection{Results}
 
We evaluated our models on four corpora, which are the movie and the Twitter datasets in Turkish and English. All of the embeddings are learnt on four corpora separately. We have used the accuracy metric since all the datasets are completely or nearly completely balanced. We performed 10-fold cross-validation for both of the datasets. We used the approximate randomisation technique to test whether our results are statistically significant. Here, we tried to predict the labels of reviews and assess the performance.
 
We obtained varying accuracies as shown in Table~\ref{SW}. ``3 feats" features are those hand-crafted features we extracted, which are the minimum, mean, and maximum polarity scores of the reviews as explained in Section 3.5. As can be seen, at least one of our methods outperforms the baseline word2vec approach for all the Turkish and English corpora, and all categories. All of our approaches performed better when we used the supervised scores, which are extracted on a review-basis, and concatenated them to word vectors. Mostly, the supervised 4-scores feature leads to the highest accuracies, since it employs the annotation information concerned with polarities on a word-basis.

\begin{table}[t!]
\caption{\label{SW} Accuracies for different feature sets fed as input into the SVM classifier in predicting the labels of reviews. The word2vec algorithm is the baseline method.}
\begin{center}
\begin{tabular}{lrrrr}
\hline
\bf{Word embedding type} & \multicolumn{2}{c}{\bf Turkish (\%)} & \multicolumn{2}{c}{\bf English (\%)} \\
&\bf{Movie}&\bf{Twitter} &\bf{Movie}&\bf{Twitter}\\
\hline
Corpus-based + SVD (U) & 76.19 & 64.38&66.54 &\textbf{87.17}\\
Dictionary-based + SVD (U) & 60.64 & 51.36&55.29 &60.00\\
Supervised 4-scores & \textbf{89.38} & \textbf{76.00}&\textbf{75.65}&72.62 \\
Concatenation of the above three & 88.12 & 73.23&73.40&73.12 \\
Corpus-based + Clustering & 52.27 & 52.73&51.02 &54.40\\
word2vec & 76.47 & 46.57&57.73&62.60 \\
\hline
Corpus-based + SVD (U) + 3-feats& 88.45 & 72.60&76.85 &\textbf{85.88}\\
Dictionary-based + SVD (U) + 3-feats & 88.64 & 71.91&76.66 &80.40\\
Supervised 4-scores + 3-feats& \textbf{90.38} & \textbf{78.00}&\textbf{77.05}&72.83 \\
Concatenation of the above three + 3 feats & 89.77 & 72.60&77.03&80.20 \\
Corpus-based + Clustering + 3-feats & 87.89 & 71.91&75.02 &74.40\\
word2vec+ 3-feats & 88.88 & 71.23&77.03&75.64 \\\hline
\end{tabular}
\end{center}
\end{table}
 
As can be seen in Table~\ref{SW}, the clustering method, in general, yields the lowest scores. We found out that the corpus - SVD metric does always perform better than the clustering method. We attribute it to that in SVD the most important singular values are taken into account. The corpus - SVD technique outperforms the word2vec algorithm for some corpora. When we do not take into account the 3-feats technique, the corpus-based SVD method yields the highest accuracies for the English Twitter dataset. We show that simple models can outperform more complex models, such as the concatenation of the three subapproaches or the word2vec algorithm. Another interesting finding is that for some cases the accuracy decreases when we utilise the polarity labels, as in the case for the English Twitter dataset.
 
Since the TDK dictionary covers most of the domain-specific vocabulary used in the movie reviews, the dictionary method performs well. However, the dictionary lacks many of the words, occurring in the tweets; therefore, its performance is not the best of all. When the TDK method is combined with the 3-feats technique, we observed a great improvement, as can be expected.
 
Success rates obtained for the movie corpus are much better than those for the Twitter dataset for most of our approaches, since tweets are, in general, much shorter and noisier. We also found out that, when choosing the \textit{p} value as 0.05, our results are statistically significant compared to the baseline approach in Turkish \cite{Ert:17}. Some of our subapproaches also produce better success rates than those sentiment analysis models employed in English \cite{Fel:17,Tan:14}. We have achieved state-of-the-art results for the sentiment classification task for both Turkish and English. As mentioned, our approaches, in general, perform best in predicting the labels of reviews when three supervised scores are additionally utilised.
 
We also employed the convolutional neural network model (CNN). However, the SVM classifier, which is a conventional machine learning algorithm, performed better. We did not include the performances of CNN for embedding types here due to the page limit of the paper.
 
As a qualitative assessment of the word representations, given some query words, we visualised the most similar words to those words using the cosine similarity metric. By assessing the similarities between a word and all the other corpus words, we can find the most akin words according to different approaches. Table~\ref{MSW} shows the most similar words to given query words. Those words which are indicative of sentiment are, in general, found to be most similar to those words of the same polarity. For example, the most similar word to \emph{muhte\c{s}em} (gorgeous) is \emph{10/10}, both of which have positive polarity. As can be seen in Table~\ref{MSW}, our corpus-based approach is more adept at capturing domain-specific features as compared to word2vec, which generally captures general semantic and syntactic characteristics, but not the sentimental ones.
 
\begin{table}[t!]
\caption{Most similar words to given queries according to our corpus-based approach and the baseline word2vec algorithm.}
\begin{center}
\begin{tabular}{lccc}
\hline
\bf{Query Word} & \bf{Corpus-based} & \bf{word2vec}\\
\hline
\multirow{2}{*}{\pbox{1.9cm}{\relax\ifvmode\centering\fi Muhte\c{s}em \emph{(Gorgeous)}}} & \multirow{2}{*}{\pbox{1.9cm}{\relax\ifvmode\centering\fi 10/10}} & \multirow{2}{*}{\pbox{1.9cm}{\relax\ifvmode\centering\fi Harika \emph{(Wonderful)}}}\\
\noalign{\vspace {.5cm}}
\multirow{2}{*}{\pbox{1.9cm}{\relax\ifvmode\centering\fi Berbat \emph{(Terrible)}}} & \multirow{2}{*}{\pbox{1.9cm}{\relax\ifvmode\centering\fi Vasat \emph{(Mediocre)}}} & \multirow{2}{*}{\pbox{1.9cm}{\relax\ifvmode\centering\fi K\"{o}t\"{u} \emph{(Bad)}}}\\
\noalign{\vspace {.5cm}}
\multirow{2}{*}{\pbox{1.9cm}{\relax\ifvmode\centering\fi Fark \emph{(Difference)}}} & \multirow{2}{*}{\pbox{1.9cm}{\relax\ifvmode\centering\fi \.{I}lgin\c{c} \\ \emph{(Interesting)}}} & \multirow{2}{*}{\pbox{1.9cm}{\relax\ifvmode\centering\fi Tespit \emph{(Finding)}}}\\
\noalign{\vspace {.5cm}}
\multirow{2}{*}{\pbox{1.9cm}{\relax\ifvmode\centering\fi K\"{o}t\"{u} \emph{(Bad)}}} & \multirow{2}{*}{\pbox{1.9cm}{\relax\ifvmode\centering\fi S{\i}k{\i}c{\i} \emph{(Boring)}}} & \multirow{2}{*}{\pbox{1.9cm}{\relax\ifvmode\centering\fi \.{I}yi \emph{(Good)}}}\\
\noalign{\vspace {.5cm}}
\multirow{2}{*}{\pbox{1.9cm}{\relax\ifvmode\centering\fi \.{I}yi \emph{(Good)}}} & \multirow{2}{*}{\pbox{1.9cm}{\relax\ifvmode\centering\fi G\"{u}zel \emph{(Beautiful)}}} & \multirow{2}{*}{\pbox{1.9cm}{\relax\ifvmode\centering\fi K\"{o}t\"{u} \emph{(Bad)}}}\\
\noalign{\vspace {.5cm}}
\multirow{2}{*}{\pbox{1.9cm}{\relax\ifvmode\centering\fi Senaryo \emph{(Script)}}} & \multirow{2}{*}{\pbox{1.9cm}{\relax\ifvmode\centering\fi Kurgu \emph{(Plot)}}} & \multirow{2}{*}{\pbox{1.9cm}{\relax\ifvmode\centering\fi Kurgu \emph{(Plot)}}}\\
\noalign{\vspace {.5cm}}
\hline
\end{tabular}
\end{center}
\label{MSW}
\end{table}
 
\section{Conclusion}
 
We have demonstrated that using word vectors that capture only semantic and syntactic characteristics may be improved by taking into account their sentimental aspects as well. Our approaches are portable to other languages and cross-domain. They can be applied to other domains and other languages than Turkish and English with minor changes.
 
Our study is one of the few ones that perform sentiment analysis in Turkish and leverages sentimental characteristics of words in generating word vectors and outperforms all the others. Any of the approaches we propose can be used independently of the others. Our approaches without using sentiment labels can be applied to other classification tasks, such as topic classification and concept mining.
 
The experiments show that even unsupervised approaches, as in the corpus-based approach, can outperform supervised approaches in classification tasks. Combining some approaches, which can compensate for what others lack, can help us build better vectors. Our word vectors are created by conventional machine learning algorithms; however, they, as in the corpus-based model, produce state-of-the-art results. Although we preferred to use a classical machine learning algorithm, which is SVM, over a neural network classifier to predict the labels of reviews, we achieved accuracies of over 90 per cent for the Turkish movie corpus and about 88 per cent for the English Twitter dataset.
 
We performed only binary sentiment classification in this study as most of the studies in the literature do. We will extend our system in future by using neutral reviews as well. We also plan to employ Turkish WordNet to enhance the generalisability of our embeddings as another future work.
 
\section*{Acknowledgments}
 
This work was supported by Boğaziçi University Research Fund Grant Number 6980D, and by Turkish Ministry of Development under the TAM Project number DPT2007K12-0610. Cem Rıfkı Aydın is supported by TÜBİTAK BIDEB 2211E.
 
\bibliographystyle{splncs}
\bibliography{paper}

\begin{thebibliography}{10}

\bibitem{Goldberg:16}
Goldberg, Y.:
\newblock A primer on neural network models for natural language processing.
\newblock Journal of Artificial Intelligence Research \textbf{1510} (2016)
  345--420

\bibitem{Mikolov:13}
Mikolov, T., Chen, K., Corrado, G., Dean, J.:
\newblock Efficient estimation of word representations in vector space.
\newblock CoRR \textbf{1301} (2013)  1--12

\bibitem{Blei:03}
Blei, D.M., Ng, A.Y., Jordan, M.I.:
\newblock Latent dirichlet allocation.
\newblock Journal of Machine Learning Research \textbf{3} (2003)  993--1022

\bibitem{Turney:10}
Turney, P.D., Pantel, P.:
\newblock From frequency to meaning: Vector space models of semantics.
\newblock Journal of Artificial Intelligence Research \textbf{37} (2010)
  141--188

\bibitem{Cortes:95}
Cortes, C., Vapnik, V.:
\newblock Support-vector networks.
\newblock Machine Learning \textbf{20} (1995)  273--297

\bibitem{Li:10}
Li, F., Huang, M., Zhu, X.:
\newblock Sentiment analysis with global topics and local dependency.
\newblock In: Proceedings of the Twenty-Fourth AAAI Conference on Artificial
  Intelligence (AAAI-10), Association for Computational Linguistics (2010)
  1371--1376

\bibitem{Lin:09}
Lin, C., He, Y.:
\newblock Joint sentiment/topic model for sentiment analysis.
\newblock In: Proceedings of the 18th ACM Conference on Information and
  Knowledge Management, ACM (2009)  375--384

\bibitem{Boyd-Graber:10}
Boyd-Graber, J., Resnik, P.:
\newblock Holistic sentiment analysis across languages: Multilingual supervised
  latent dirichlet allocation.
\newblock In: Proceedings of the 2010 Conference on Empirical Methods in
  Natural Language Processing (EMNLP), Association for Computational
  Linguistics (2010)  45--55

\bibitem{Maas:11}
Maas, A.L., Daly, R.E., Pham, P.T., Huang, D., Ng, A.Y., Potts, C.:
\newblock Learning word vectors for sentiment analysis.
\newblock In: Proceedings of the 49th Annual Meeting of the Association for
  Computational Linguistics: Human Language Technologies - Volume 1,
  Association for Computational Linguistics (2011)  142--150

\bibitem{Hamilton:16}
Hamilton, W.L., Clark, K., Leskovec, J., Jurafsky, D.:
\newblock Inducing domain-specific sentiment lexicons from unlabeled corpora.
\newblock In: Proceedings of the 2016 Conference on Empirical Methods in
  Natural Language Processing (EMNLP), Association for Computational
  Linguistics (2016)  595--605

\bibitem{Ert:17}
Ertugrul, A.M., {\"{O}}nal, I., Acart{\"{u}}rk, C.:
\newblock Does the strength of sentiment matter? {A} regression based approach
  on turkish social media.
\newblock In: Natural Language Processing and Information Systems - 22nd
  International Conference on Applications of Natural Language to Information
  Systems, {NLDB} 2017, Li{\`{e}}ge, Belgium, June 21-23, 2017, Proceedings.
  (2017)  149--155

\bibitem{Fel:17}
Felbo, B., Mislove, A., S{\o}gaard, A., Rahwan, I., Lehmann, S.:
\newblock Using millions of emoji occurrences to learn any-domain
  representations for detecting sentiment, emotion and sarcasm.
\newblock In: {EMNLP}, Association for Computational Linguistics (2017)
  1615--1625

\bibitem{Tan:14}
Tang, D., Wei, F., Qin, B., Liu, T., Zhou, M.:
\newblock Coooolll: {A} deep learning system for twitter sentiment
  classification.
\newblock In: Proceedings of the 8th International Workshop on Semantic
  Evaluation, SemEval@COLING 2014, Dublin, Ireland, August 23-24, 2014. (2014)
  208--212

\bibitem{Bac:10}
Baccianella, S., Esuli, A., Sebastiani, F.:
\newblock Sentiwordnet 3.0: An enhanced lexical resource for sentiment analysis
  and opinion mining.
\newblock In Calzolari, N., Choukri, K., Maegaard, B., Mariani, J., Odijk, J.,
  Piperidis, S., Rosner, M., Tapias, D., eds.: LREC, European Language
  Resources Association (2010)

\bibitem{Go:09}
Go, A., Bhayani, R., Huang, L.:
\newblock Twitter sentiment classification using distant supervision.
\newblock Processing (2009)  1--6

\bibitem{Akin:07}
Ak{\i}n, A.A., Ak{\i}n, M.D.:
\newblock Zemberek, an open source nlp framework for turkic languages.
\newblock Structure \textbf{10} (2007)  1--5

\bibitem{Sak:07}
Sak, H., G\"{u}ng\"{o}r, T., Sara\c{c}lar, M.:
\newblock Morphological disambiguation of turkish text with perceptron
  algorithm.
\newblock In: Proceedings of the 8th International Conference on Computational
  Linguistics and Intelligent Text Processing (CICLing 2017), CICLing Press
  (2007)  107--118

\bibitem{Sak:08}
Sak, H., G\"{u}ng\"{o}r, T., Sara\c{c}lar, M.:
\newblock Turkish language resources: Morphological parser, morphological
  disambiguator and web corpus.
\newblock In: Proceedings of the 6th International Conference on Advances in
  Natural Language Processing (GoTAL '08), Springer-Verlag (2008)  417--427

\bibitem{Rad:10}
{\v R}eh{\r u}{\v r}ek, R., Sojka, P.:
\newblock {Software Framework for Topic Modelling with Large Corpora}.
\newblock In: {Proceedings of the LREC 2010 Workshop on New Challenges for NLP
  Frameworks}, Valletta, Malta, ELRA (2010)  45--50
  \url{http://is.muni.cz/publication/884893/en}.

\end{thebibliography}
\end{document}